\documentclass[a4paper, 10pt, conference]{ieeeconf}      
\usepackage{FG2019}

\FGfinalcopy 

\IEEEoverridecommandlockouts                              
\usepackage{graphicx} 
\usepackage{amsmath} 
\usepackage{amssymb}  

\usepackage{algorithm}
\usepackage[noend]{algpseudocode}
\usepackage{multirow}
\usepackage{url}

\usepackage{soul}

\def\FGPaperID{} 

\title{\LARGE \bf
	Adversarial-based neural networks for affect estimations in the wild 
}

\author{\parbox{16cm}{\centering
		{\large Decky Aspandi$^{1,2}$, Adria Mallol-Ragolta$^2$,  Bj\"orn W.\ Schuller$^{2,3}$, and Xavier Binefa$^1$}\\
		{\normalsize
			$^{1}$ Department of Information and Communication Technologies,  Pompeu Fabra University, Barcelona, Spain\\[1.5pt]
			$^{2}$ Chair of Embedded Intelligence for Health Care \& Wellbeing, University of Augsburg, Germany\\
			$^{3}$ GLAM -- Group on Language, Audio \& Music, Imperial College London, UK}}
}

\begin{document}
	
	\ifFGfinal
	\thispagestyle{empty}
	\pagestyle{empty}
	\else
	\author{Anonymous FG 2019 submission\\ Paper ID \FGPaperID \\}
	\pagestyle{plain}
	\fi
	\maketitle

	\begin{abstract}
There is a growing interest in affective computing research nowadays given its crucial role in bridging humans with computers. This progress has been recently accelerated due to the emergence of bigger data. One recent advance in this field is the use of adversarial learning to improve model learning through augmented samples. However, the use of latent features, which is feasible through adversarial learning, is not largely explored, yet. This technique may also improve the performance of affective models, as analogously demonstrated in related fields, such as computer vision. To expand this analysis, in this work, we explore the use of latent features through our proposed adversarial-based networks for valence and arousal recognition in the wild. Specifically, our models operate by aggregating several modalities to our discriminator, which  is further conditioned to the extracted latent features by the generator. Our experiments on the recently released SEWA dataset suggest the progressive improvements of our results. Finally, we show our competitive results on the Affective Behavior Analysis in-the-Wild (ABAW) challenge dataset.

	\end{abstract}
	
	\section{INTRODUCTION}\label{sec:introduction}
	

Affective computing has recently attracted the attention of the research community, due to its applications in multiple and diverse areas, including education~\cite{e_learning} or healthcare~\cite{Liu}, among others. Furthermore, the growing availability of affect-related datasets, such as SEWA~\cite{sewa} and the recently introduced Aff-Wild2~\cite{kollias2019expression}, enable the rapid development of deep learning-based techniques, which currently hold the state of the art~\cite{afew,sewa,kollias2019deep}. 

In computer vision tasks, such as natural image generation\cite{GAN} and image classification\cite{semigan}, adversarial learning techniques from the family of generative models have been extensively investigated~\cite{GAN,semigan,stargan}. This learning technique enables rapid progress, not only to create additional data, but also to improve the performance of predictive models. Nevertheless, in the context of affective computing-related applications, this technique is still young and confined to its usage for data augmentation purposes~\cite{han2019adversarial}.

To expand the investigation of generative models in the field of affective computing, we investigate the use of latent features that are extracted in adversarial manners to improve the predictive capabilities of our model estimations. Specifically, we extract the visual latent features of the generator, which are then used to condition the discriminator on its estimations. Furthermore, we also aggregate the audio modality during training. We later show in our experiments on the SEWA~\cite{sewa} and Aff-Wild2~\cite{kollias2019expression} datasets the benefits of our proposed approach with our competitive results. Specifically, the contributions of this work are: 
\begin{enumerate}
	\item We are the first to introduce the utilisation of latent features arranged in an adversarial way to improve affect-related model estimates. 
	\item We show the progressive improvements on our proposed works on the SEWA and Aff-Wild2 datasets and achieve competitive results on both datasets.
\end{enumerate}

\section{RELATED WORKS}\label{sec:relatedWorks}

\begin{figure*}[t!]
	\begin{center}
		\includegraphics[width=1\linewidth]
		{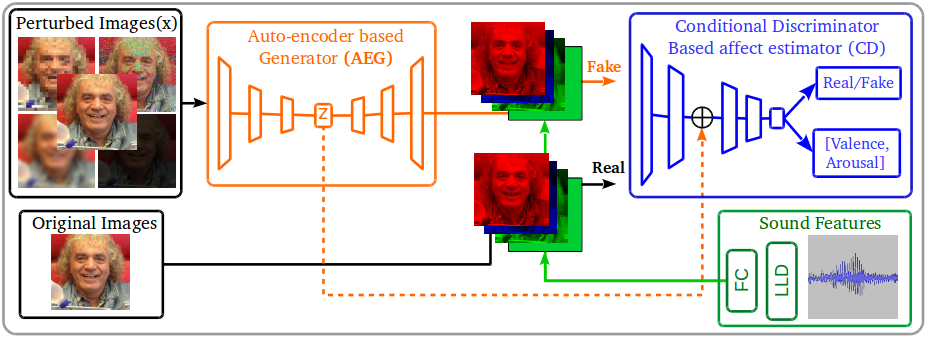}
	\end{center}
	\caption{Complete architecture of our proposed models which incorporate two main networks: first is an Auto-Encoder-based Generator (AEG) which denoises the image and creates robust latent features. Second is a  Conditional Discriminator-based affect estimator (CD) that aggregates both sounds and image input which is conditioned by latent features from the CD to estimate both real/fake and valence/arousal values.  
	}
	\label{fig:archi}
\end{figure*}
Early approaches on automatic affect estimations involved the use of classical machine learning techniques with some degree of success. Several techniques explored include linear regression~\cite{linearReg}, partial least square regression~\cite{leastSqu}, and support vector machines~\cite{svr}. Furthermore, given the number of available modalities (e.\,g.\, video, audio,  and bio-signals), several fusion techniques were also introduced to improve  affect-related estimates. Different examples of these methods include early, late, model, and output-associative fusion~\cite{fusion}. Diverse affective information has been progressed, starting from Action Units detection, emotion detection, to more recently continuous valence and arousal estimation\cite{afew,sewa}.

Current progress relates to the emergence of big data that creates the opportunity to introduce large scale datasets in many fields, including affective computing. Examples of these datasets are SEMAINE~\cite{semaine}, AVEC~\cite{avec}, AFEW~\cite{afew}, RECOLA~\cite{recola}, SEWA~\cite{sewa}, and the recently introduced Aff-Wild2 dataset~\cite{kollias2019expression,kollias2018multi,zafeiriou2017aff,kollias2017recognition}. These datasets enable the development of powerful deep learning models that improve the accuracy of current state of the art~\cite{afew,kollias2018aff,kollias2019deep}. The investigations of deep learning-based techniques include the introduction of Convolutional Neural Networks (CNN)~\cite{cnn}, incorporation of Recurrent Neural Network (RNN)~\cite{sewa, rnn}, and recently the fusion with Tensor-based methods~\cite{mitenkova2019valence}. 

Adversarial learning~\cite{GAN} as a generative approach has been intensively studied in other machine learning research, especially in computer vision~\cite{stargan}. Given its potential, this method has also been explored in the field of affective computing, usually to augment the training data available for training~\cite{han2019adversarial}. However, there is another aspect of generative models that is largely unexplored in this field, which is the use of latent features to improve the models estimations, as shown in previous works from other fields, such as computer vision~\cite{Trumble_2018_ECCV}, machine learning~\cite{aeML}, and bio-signal analysis~\cite{tang2017multimodal, comas2019endtoend}. This inspired us to investigate the use of adversarial learning to improve our proposed models' performance through extracted latent features.

\section{Adversarial latent-based networks} \label{sec:method}

We build our model based on the Star-GAN network\cite{stargan}, with architectural modifications to allow the extraction of latent features and use of the audio features. Figure \ref{fig:archi} shows the overview of our proposed network.  Our model operates by aggregating two main modalities: facial and audio features. There are two main sub-networks involved in our overall networks as already outlined above: the Auto-Encoder-based Generator(AEG), and the Conditional Discriminator-based affect estimator (CD)~\cite{sgan}. The main role of the AEG is to produce cleaned images from noisy images to fool the discriminator, while simultaneously extracting robust latent features.  On the other hand, the CD tries to recognise the fake images created by the AEG, and, at the same time, estimates the actual valence and arousal values. We train the AEG and CD in an adversarial way as below: 
\begin{equation}
\begin{split}
\mathcal{L}_{adv} = & \thinspace {\mathbb{E}}_{x} \left[ \log{{CD}_{adv}(x)} \right]  \> \>  +   \\
& \thinspace {\mathbb{E}}_{x}[\log{(1 - {CD}_{adv}(AEG(\hat{x})))}],
\end{split}
\label{eq1}
\end{equation}
where ${x}$ corresponds to the noisy image and $\hat{x}$ is the cleaned input image approximated by the AEG. We use similar noise introduction methods as in \cite{FADeNN}, which consist of four different types of artifacts: Gaussian blurring, Gaussian noise, image downsampling, and colour scaling.



\subsection{Auto-encoder-based generator}
Given the noisy input image, $x$, the AEG will approximate the cleaned version of the input image, $\hat{x}$. This is done by utilising coupled mirrored convolutions and deconvolutions with intermediate 2D bottleneck latent kernels; i.\,e.\, without skip connections. This scheme enforces the AEG to create latent robust features in order to effectively clean the input image. To improve the denoising and reconstruction process, we use the cycle loss \cite{stargan,kim2017learning} defined below : 
\begin{equation}
\mathcal{L}_{rec} = {\mathbb{E}}_{x} [{||x - AEG(AEG(\hat{x}))||} ].
\label{eq4}
\end{equation}

\subsection{Conditional discriminator-based affect estimator}
The CD employs both facial and audio features to identify the real/fake status of the current input and the corresponding valence and arousal values of $\hat{\theta}$. The facial features correspond to the cleaned image (denoised or reconstructed from the generator and the corresponding latent features of $z$. From the audio modality, we use the low-level descriptors (LLDs) of the \textsc{eGeMAPS} feature set\cite{eyben16-TGM} (cf.\,Section \ref{sec:sfe}). Both latent and audio features are combined through late fusion \cite{gunes2005affect,snoek2005early,fusion} alongside the main RGB input images. Specifically, the audio features are merged by feeding them into a 1D fully connected layer to enlarge its dimension and converting it to a single 2D kernel, which is then concatenated with the denoised image. The latent features are combined in middle pipelines of the CD by concatenating them with intermediate kernels. 

To detect both real and fake status and estimate valence and arousal values, we add another classifier\cite{semigan} on top of the main classifier, which consists of a 2x2 pixels layer\cite{patchGAN}. In the adversarial training, the CD will be optimised using real ($r$) and fake ($f$) images to minimise the affect loss ($\mathcal{L}_{afc}$) that judges the accuracy of the estimated valence and arousal values (cf.\ Equation \ref{afc}). The corresponding loss of training the CD for both real ($\mathcal{L}_{va}^{r}$) and false examples ($\mathcal{L}_{va}^{f}$) can be seen below : 



\begin{equation}
\mathcal{L}_{va}^{r} = {\mathbb{E}}_{x, \theta}[-\mathcal{L}_{afc}(\theta'|x)],
\label{eq2}
\end{equation}
\begin{equation}
\mathcal{L}_{va}^{f} ={\mathbb{E}}_{x, \theta}[-\mathcal{L}_{afc}({D}(\theta|G(x)))],
\label{eq3}
\end{equation}
where $\hat{\theta}$ is the ground truth valence/arousal value, and the affect loss, $\mathcal{L}_{afc}$, corresponds to the amalgamations of multiple affect metrics: Mean Square Error(MSE) (Eq.\ \ref{mse}), Correlation(COR) (Eq.\ \ref{cor}), and Concordance Correlation Coefficients (CCC) (Eq.\ \ref{ccc}), \cite{afew,kollias2019expression} : 

\begin{gather}
\label{afc}
\mathcal{L}_{afc} = \sum_{i = 1}^{N} \frac{n_i}{N} (\mathcal{L}_{MSE} + \mathcal{L}_{COR} + \mathcal{L}_{CCC}) \\
\label{mse}
\mathcal{L}_{MSE} = \sqrt{\frac{1}{n} \sum_{i=1}^{n} (\hat{\theta_i},\theta_i)}, \\ 
\label{cor}
\mathcal{L}_{COR} = \frac{\mathbb{E}[(\hat{\theta}-\mu_{\theta}) - (\theta-\mu_{\theta})]}{\sigma_{\hat{\theta}}\sigma_{\theta}}  \\ 
\label{ccc}
\mathcal{L}_{CCC} =  2x\frac{\mathbb{E}[(\hat{\theta}-\mu_{\theta}) - (\theta-\mu_{\theta})]}{\sigma_{\hat{\theta}}^2+\sigma_{\theta}^2},
\end{gather}
where $n_i$ is the total number of instances of discrete valence/arousal class $i$, and $N$ is the normalisation factor\cite{NoiseRedFamily} for the total valence/arousal class. This normalisation factor is crucial given considerably unbalanced class instance on the Aff-Wild2 Challenge\cite{kollias2020analysing}.




\subsection{Overall objective} 
Finally, the overall objective functions to train both AEG and CD are expressed as follows: 
\begin{equation}
\mathcal{L}_{D} =  - \mathcal {L}_{adv} +  {\lambda}_{afc}\thinspace\mathcal{L}_{afc}^{r}, 
\end{equation}
\begin{equation}
\mathcal{L}_{G} =    \mathcal {L}_{adv} +  {\lambda}_{afc}\thinspace\mathcal{L}_{afc}^{f} + 
{\lambda}_{rec}\thinspace\mathcal{L}_{rec},
\end{equation}
in which ${\lambda}_{afc}$ and ${\lambda}_{rec}$ are the regulariser parameters for affect estimations and reconstruction loss. 




\subsection{Audio feature extraction}\label{sec:sfe}
One of the first challenges when combining audio and video signals is the difference in term of sampling rates between both modalities. To overcome this issue, we first generate audio frames from the original audio signal by selecting the portions of the audio signal corresponding to one frame of video. We then enlarge the audio frame with the samples corresponding to the previous and future video frame to ensure information overlap between consecutive audio frames. We finally extract the LLDs of the \textsc{eGeMAPS}~~\cite{eyben16-TGM} feature set using \textsc{openSMILE}~~\cite{openSMILE}, and concatenate the first two sets of LLDs for further analysis. These LLDs are extracted from windows of 0.060 seconds with a step size of 0.010 seconds. Selecting the first two sets of LLDs only, we ensure the same dimensionality of the audio features in spite of videos recorded at different sampling rates.

\subsection{Model training}
In the challenge dataset, we only utilise the training subset to obtain our validation results. We exploit the full available data (training and validation) to train our final models submitted to the challenge organisers. Specifically, we used the crop-aligned samples provided by the organisers as facial features, in addition to the audio signals from the available videos. 

To provide more actual comparisons of our results to the other state of the arts, we also include the experiments on the Sentiment Analysis in the Wild dataset (SEWA)\cite{sewa}. To obtain our results, we followed original person-independence protocols, and apply similar feature extraction techniques as those employed for this challenge. Furthermore, we also use the external tracker of \cite{CRCT} to refine the given bounding box. 

For both datasets, we trained our model progressively to allow us to analyse the impact of each proposed step. We first train both of generator and discriminator together, and proceed by adding the extracted latent $z$ features alongside the audio features. Our models were trained using an NVIDIA Titan X GPU and it took approximately two days to converge. The source code of our models is available at our github page\footnote{https://github.com/deckyal/ALN}
\section{EXPERIMENTS}\label{sec:results}
In this section, we describe our results on the recently introduced Aff-Wild2 Challenge\cite{kollias2020analysing} and SEWA\cite{sewa} datasets to confirm the advantages of each of our proposed approaches.  
\begin{itemize}
    \item The Aff-Wild2 challenge dataset is being published as part of the first ABAW 2020 competitions\cite{kollias2020analysing}. This competition consists of three main challenges : valence-arousal, basic expression and eight action units. Aff-wild2 is considered to be the current, largest affect in the wild dataset with more than 558 videos and 458 total number of subjects. Specific on the valence and arousal challenge, there are 545 annotated videos with 2.786.201 frames which is split into three subsets : 346 videos of training, 68 videos of validation and 131 videos of test.
    \item The SEWA dataset\cite{sewa} is a recently published affect dataset which consists of video and audio recording involving 398 subjects from multiple cultures. This dataset is split into 538 sequences with various meta-data (e.g. subject id, culture etc) are available alongside the actual affect ground truth of valence/arousal and liking/disliking.  
\end{itemize}


We use MSE, COR and CCC metrics to evaluate the quality of each affect estimations\cite{sewa,avec,mitenkova2019valence,kollias2020analysing}. That on the Aff-Wild2 dataset, we compared our results on the validation stage against the baseline provided by the organizers \cite{kollias2020analysing}. While for the SEWA dataset, we report our results from original five cross validation settings \cite{sewa} and compared them with the respective baseline\cite{sewa} and recent state of the art of \cite{mitenkova2019valence}



Tables \ref{tab:sewa} and \ref{tab:challenge} provide our results on ABAW Challenge and SEWA, respectively. Method Disc corresponds to our results utilizing only plain Discriminator (CD) trained using standard $\ell^2$ loss. Method AEG-CD means that our model uses adversarial training for both of AEG and CD. Lastly, the AEG-CD-ZS shows the results of our previous model trained with the inclusion of both latent features $z$ from AEG and the mapped audio features.



\begin{table}[t!]
	\resizebox{\columnwidth}{!}{%
		\begin{tabular}{lcccccc}
			\hline
			\multirow{2}{*}{Methods} & \multicolumn{2}{c}{MSE} & \multicolumn{2}{c}{COR} & \multicolumn{2}{c}{CCC} \\
			& Val&Aro & Val&Aro & Val&Aro  \\
			\hline
			Baseline~\cite{kollias2020analysing} & - & - & - & - & \textbf{0.14} & 0.24 \\
			Disc & 0.44 & 0.30 & 0.07 & 0.19 & 0.07 & 0.20 \\
			AEG-CD & 0.42 & 0.28 & 0.10 & 0.22 & 0.08 & 0.22 \\
			AEG-CD-SZ & 0.42 & 0.28 & 0.11 & 0.29 & 0.10 & \textbf{0.26} \\
			\hline
		\end{tabular}%
	}
	\caption{Experiments on ABAW Challenge dataset}
	\label{tab:challenge}
\end{table}

\begin{table}[t!]
	\resizebox{\columnwidth}{!}{%
		\begin{tabular}{lcccccc}
			\hline
			\multirow{2}{*}{Methods} & \multicolumn{2}{c}{MSE} & \multicolumn{2}{c}{COR} & \multicolumn{2}{c}{CCC} \\
			& Val&Aro & Val&Aro & Val&Aro  \\
			
			\hline
			Baseline~\cite{sewa}                & - & - & 0.322 & 0.4 & 0.195  & 0.427 \\
			Tensor~\cite{mitenkova2019valence}  & 0.334 & 0.380 & \textbf{0.503}  & 0.439 & \textbf{0.469} & 0.392 \\
			Disc                            & 0.336 & 0.399 & 0.395 & 0.457 & 0.349 & 0.379 \\
			AEG-CD                        & 0.329 & 0.394 & 0.429 & 0.467 & 0.380 & 0.429 \\
			AEG-CD-SZ                          & \textbf{0.323} & \textbf{0.350} & 0.442 & \textbf{0.478} & 0.405 & \textbf{0.430} \\
			\hline
		\end{tabular}%
	}\caption{Experiments on SEWA dataset}
	\label{tab:sewa}
\end{table}

\begin{figure}[t!]
	\begin{center}
		\includegraphics[width=.8\linewidth]
		{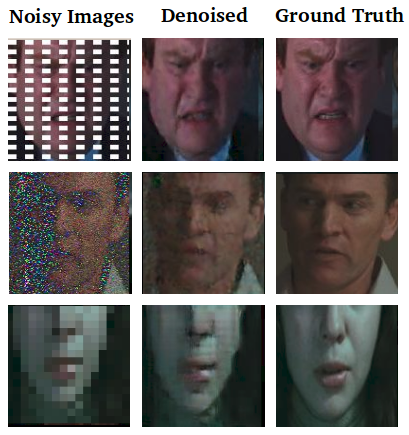}
	\end{center}
	\caption{Example of denoised images. As we can see, the denoised image is quite cleaned and almost resembles the ground truth}
	\label{fig:den}
\end{figure}

Based on these quantitative results, we can see that our models are able to produce quite competitive results, with high accuracy on the arousal dimension. Specifically, we observe quite high accuracy obtained by our discriminator (Disc) which is able to outperform the current baseline results of SEWA, albeit it is still lower on the challenge dataset. This maybe attributed to limitations of using standard loss, which does not incorporate more objective metrics that adversarial learning offers\cite{stargan, semigan, Gan2015}. This is confirmed by the better accuracy obtained by using adversarial learning. Another potential reason can be attributed to the generated images that may reduce the available noises on the input images. An example of denoised images can be seen in Figure \ref{fig:den}. There, we can see that the denoised input image is quite cleaned and this may also be indicative of a successful generator training in creating the robust latent features. We finally can see that incorporating both of the latent and audio features improves the overall accuracy of our results, surpasses both of the baselines on the affect dataset, and the state of the art of the SEWA dataset, especially on the arousal dimension. This result highlights the benefit of incorporating such features.


\section{CONCLUSION} \label{sec:conclusion}
In this paper, we presented the first investigation using latent features extracted through adversarial learning in Affective Computing domain. Specifically, we performed progressive training on our generator to extract robust features given noisy inputs paired with a discriminator through adversarial learning. Then, we employed a conditional discriminator to aggregate several modality's inputs to achieve our affect estimations. We tested the performance of our models on two datasets: Aff-Wild2 and SEWA. In our experiments, we observed progressive improvements made by each of our approaches utlimately leading to competitive results on the ABAW challenge dataset. In the future, we seek to incorporate temporal modelling to further increase the accuracy of the proposed models. 



\section{ACKNOWLEDGMENTS}
This work is partly supported by the Spanish Ministry of Economy and Competitiveness under project grant TIN2017-90124-P, the Ramon y Cajal programme, the Maria de Maeztu Units of Excellence Programme (MDM-2015-0502),  and the donation bahi2018-19 to the CMTech at UPF. Further funding has been received from the European Union's Horizon 2020 research and innovation programme under grant agreement No.\ 826506 (sustAGE).



{\small
	\bibliographystyle{ieee}
	\bibliography{Tracking}
}

\end{document}